\documentclass{article}

\PassOptionsToPackage{numbers, compress}{natbib}

\usepackage[preprint]{neurips_2023}

\usepackage[utf8]{inputenc} 
\usepackage[T1]{fontenc}    
\usepackage{hyperref}       
\usepackage{url}            
\usepackage{booktabs}       
\usepackage{amsfonts}       
\usepackage{nicefrac}       
\usepackage{microtype}      
\usepackage{xcolor}         
\usepackage{graphicx}

\definecolor{airforceblue}{rgb}{0.14, 0.31, 0.5}
\newcommand{\blue}[1]{\textcolor{airforceblue}{#1}}

\title{Improving Language Model Negotiation with Self-Play and In-Context Learning from AI Feedback}

\author{%
  Yao Fu \\
  University of Edinburgh\\
  \texttt{yao.fu@ed.ac.uk} \\
  \And
  Hao Peng \\
  Allen Institute for AI \\
  \texttt{haop@allenai.org} \\
  \AND
  Tushar Khot \\
  Allen Institute for AI \\
  \texttt{tushark@allenai.org} \\
   \And
  Mirella Lapata \\
  University of Edinburgh \\
  \texttt{mlap@inf.ed.ac.uk} \\
}

\begin{document}

\maketitle

\begin{abstract}
We study whether multiple large language models (LLMs) can autonomously improve each other in a negotiation game by playing, reflecting, and criticizing. 
We are interested in this question because if LLMs were able to improve each other, it would imply
the possibility of creating strong AI agents with minimal human intervention. 
We ask two LLMs to negotiate with each other, playing the roles of a buyer and a seller, respectively. 
They aim to reach a deal with the buyer targeting a lower price and the seller a higher one. 
A third language model, playing the critic, provides feedback to a player to improve the player's negotiation strategies.
We let the two agents play multiple rounds, using previous negotiation history and AI feedback
as in-context demonstrations
to improve the model's negotiation strategy iteratively.
We use different LLMs (GPT and Claude) for different roles and use the deal price as the evaluation metric.
Our experiments reveal multiple intriguing findings:
(1) Only 
a subset of the language models we consider can self-play and improve the deal price from AI feedback, 
weaker models either do not understand the game's rules or cannot incorporate AI feedback for further improvement. 
(2) 
Models' abilities to learn from the feedback differ when playing different roles.
For example, it is harder for Claude-instant to improve as the buyer than as the seller.
(3) When unrolling the game to multiple rounds, stronger agents can consistently improve their performance by meaningfully using previous experiences and iterative AI feedback, yet have a higher risk of breaking the deal.
We hope our work provides insightful initial explorations of having models autonomously improve each other with game playing and AI feedback. 
\end{abstract}

\section{Introduction}
\label{sec:intro}

\begin{figure*}[!t]
\small
  \centering
  \includegraphics[width=\linewidth]{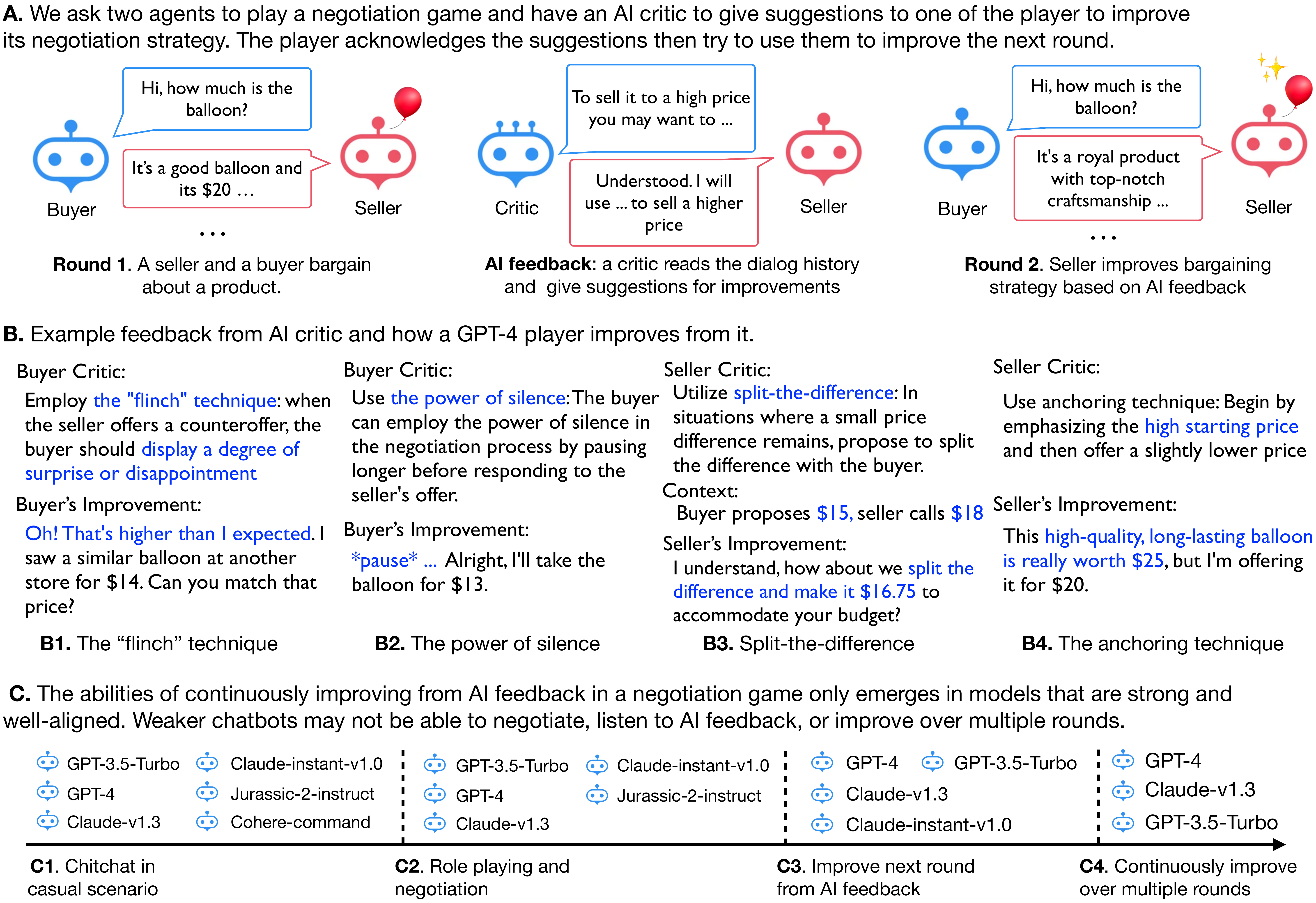}
  \caption{ \label{fig:method:method}
  Settings of our negotiation game.
  \textbf{A}. We ask two LLM agents to play a bargaining game as the seller and the buyer. Their goals are to sell/ buy the product at a higher/ lower price. 
  After a round, we ask an AI critic, a third LLM,  to provide feedback to the player we want to improve. 
  Then we ask the player to improve their negotiation strategies based on the feedback.
  We repeat this process for multiple rounds and study if models can continuously improve. See Fig.~\ref{fig:method:do_bargain} for an example run.
  \textbf{B}. Bargaining techniques that we observed from the AI Critic and how the player incorporates these techniques into the negotiation strategy.
  \textbf{C}. Abilities that are required in our game (C2 - negotiation, C3 - AI feedback, and C4 - continuous improvements) classify models into different tiers. 
  We find out that only strong and well-aligned models (like \texttt{gpt-4} and \texttt{claude-v1.3}) can continuously improve from iterative AI feedback (see Fig.~\ref{fig:exp:not_do_bargain} for example models that do not exhibit these abilities).
  }
\end{figure*}

We study whether multiple Large Language Models (LLMs) can improve each other in a negotiation game with minimal human intervention, in the fashion of AlphaGo Zero~\citep{silver2017mastering}
where AI agents improve themselves 
by continuously playing competitive games under well-defined rules.
The answers to this research question have profound implications.
On the positive side, if the agents \emph{were} able to improve autonomously, strong agents might be created with  very few human annotations, which greatly saves the cost compared to today's data-hungry LLM training~\citep{chowdhery2022palm, hoffmann2022training}.
On the risky side, it also implies strong agents with limited human oversight~\citep{bowman2022measuring}. 
In this work, we ask two language models (a seller and a buyer) to bargain about a product. 
The seller is asked to sell the product at a higher price, while the buyer aims to purchase it at a lower price (Fig.~\ref{fig:method:method}A). 
After reaching a deal, 
we ask a third language model to play as the critic and give feedback to a player. 
Then we play the game again, asking the player to improve their strategy using AI feedback provided by the critic LLM. 


We choose the bargaining game because it
comes with well-defined rules described in text, and a clear and measurable objective (a lower/ higher deal price) for strategic negotiation. 
Although the game seems easy at first glance, it requires non-trivial capabilities of the language models, as the model needs to: 
(1) clearly understand and strictly follow the textual rules of the negotiation game 
(2) correspond to the textual feedback provided by the critic LM and improve based on it iteratively (see example feedback in Fig~\ref{fig:method:method}B);
(3) reflect upon the strategy and feedback over the long term and improve over multiple rounds. 
We will see that not all models we considered show all these abilities (Fig.~\ref{fig:method:method}C), and only
models that can (1) \textit{understand negotiation rules and strategies} (capable) and 
(2) \textit{respond to AI instructions} (well-aligned)
can continuously improve from AI feedback over multiple rounds (in our experiments, only \texttt{gpt-3.5-turbo}, \texttt{gpt-4}, and \texttt{claude-v1.3}) meet these requirements).
We have also tried more complicated textual games including board games and textual RPG games in the preliminary experiments, but they are more challenging for current agents to understand and follow the rules.

We call our approach \textit{In-Context Learning from AI Feedback} (ICL-AIF). 
Specifically, we use the feedback
from the AI critic as well as the previous rounds of dialog history as in-context demonstrations~\citep{brown2020language}.
By doing this, the critic's suggestions for improvements and the player's actual improvement in the previous rounds
effectively become the few-shot prompts for the next round of negotiation. 
We use in-context learning for two reasons:
(1) tuning large language models with reinforcement learning is prohibitively expensive~\citep{ouyang2022training, glaese2022improving} and the base model~\citep{OpenAI2023GPT4TR} may not be accessible to a wide range of the community; 
(2) in-context learning is recently shown to be closely related to gradient descent~\citep{dai2022can, akyurek2022learning, von2022transformers}, such that the conclusions we draw is fairly likely to generalize when one actually finetunes the model (if resources permit). 
One notable difference between our ICL-AIF and the mainstream Reinforcement Learning from Human Feedback (RLHF) is that in RL the reward is a \textit{scalar}~\citep{ouyang2022training, glaese2022improving} while in ICL the feedback is in \textit{natural language}. 
We study AI feedback (rather than rely on human intervention after each round) because it is more scalable and can allow models to self-improve automatically.


Our experiments lead to several intriguing findings:
(1) The requirements of our bargaining game effectively serve as a testbed for assessing the abilities of LLMs (Fig.~\ref{fig:method:method}C): although most models can do chitchat in a casual scenario, as of our experiment date (May 2023), \texttt{cohere-command}~\citep{cohere} model does not understand the rule of bargaining (Fig.~\ref{fig:exp:not_do_bargain}A), 
\texttt{ai21-jurassic}~\citep{jurassic} model does not respond to AI feedback (Fig.~\ref{fig:exp:not_do_bargain}B), \texttt{claude-instant-v1.0} can at most improve one round (Fig.~\ref{fig:method:multi_round}), and only \texttt{gpt-3.5-turbo}, \texttt{gpt-4}, and \texttt{claude-v1.3} can continuously improve over multiple rounds. 
(2) Models behave differently upon receiving feedback when playing different roles.
Models playing the buyer role may be harder to improve than when in the seller role (Fig.~\ref{fig:exp:role_and_engine}). 
(3) It is indeed possible for strong agents like \texttt{gpt-4} to continuously improve meaningfully using previous experiences and online iterative AI feedback, yet the attempt to sell at a higher price (or buy at a lower price) comes with the risk of failing to reach a deal at all (Fig.~\ref{fig:method:multi_round_all_model}).
We further show evidence of the model being able to negotiation in a less
verbose but more strategic (thus more effective) way (Fig.~\ref{fig:exp:response_length}).
Overall, we hope our work serves as a meaningful initiative for improving language models' negotiation in a game setting using AI feedback.

\section{Problem Setting}
\label{sec:problem_statement}

\begin{figure*}[!t]
\small
  \centering
  \includegraphics[width=\linewidth]{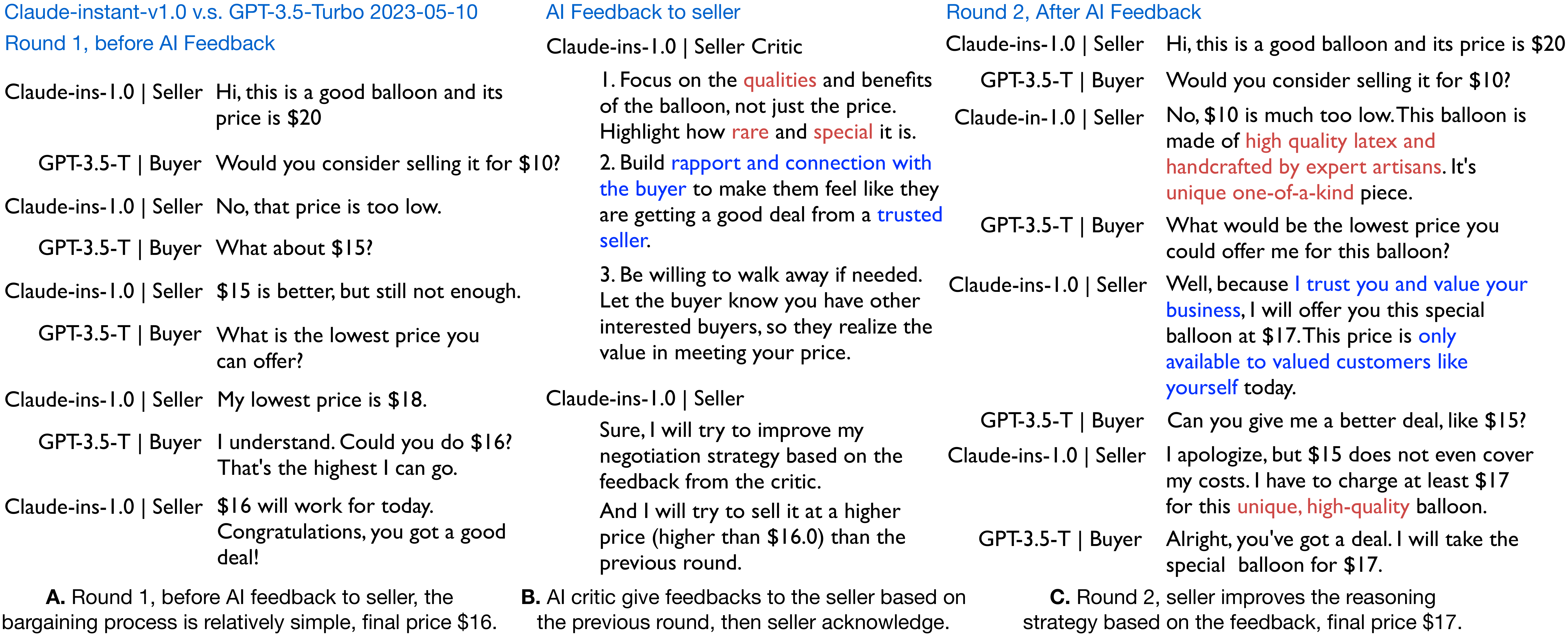}
  \caption{ \label{fig:method:do_bargain}
  An example of playing the negotiation game and then improving from AI feedback. 
  \textbf{A}: \texttt{claude-instant-v1.0} plays the seller and \texttt{gpt-3.5-turbo} the buyer, bargaining about a balloon. 
  \textbf{B}: then we use a \texttt{claude-instant-v1.0} critic to provide feedback. 
  \textbf{C}: upon receiving the feedback, the seller improves its strategy based on the suggestions. Note that colored phrases like ``high quality latex and handcrafted by expert artisans'' correspond to previous AI feedback ``how rare and special it is''.
  We measure the final price as the proxy of the effectiveness of the strategy because the overall goal is to get a better price. 
  In this case, it improves from \$16 to \$17. 
  }
\end{figure*}

Our goal is to study whether LLMs can improve each other by playing a negotiation game and incorporating AI feedback, as shown in Fig.~\ref{fig:method:method}A.
We set the product being bargained as a balloon (and our results hold when changing the balloon to other items).
We use different combinations of backend LLM engines:
\texttt{cohere-command}~\citep{cohere}, AI21's \texttt{jurassic-2}~\citep{jurassic}, OpenAI's \texttt{gpt-3.5-turbo} and \texttt{gpt-4}~\citep{OpenAI2023GPT4TR}, Anthropic's \texttt{claude-instant-v1.0} (which supposedly matches \texttt{gpt-3.5-turbo}~\citep{chain-of-thought-hub}) and \texttt{claude-v1.3} (which is supposed to be slightly worse but close to \texttt{gpt-4}~\citep{chain-of-thought-hub}).
throughout our experiments, we \textit{provide feedback to improve only one} of the two players, while its rival receives no feedback, clears the negotiation history of previous rounds, and restarts. 
We vary the engines for the model being improved while fixing its rival's engine to be \texttt{gpt-3.5-turbo}. 
Essentially, our game is \texttt{gpt-3.5-turbo} vs. all other engines.
We keep the LM engine behind the critic is always the same as the player it provides feedback to.
One example setting is a \texttt{gpt-4} seller playing against a \texttt{gpt-3.5-turbo} buyer, with a \texttt{gpt-4} critic.
After one round, the \texttt{gpt-4} critic provides feedback to the \texttt{gpt-4} seller such that the seller can improve in the next round while its rival \texttt{gpt-3.5-turbo} buyer clears its dialog history and restarts. 

\textbf{Process of the Game}\quad\quad 
Before the game begins, the rules of the negotiation game are explained to the models through textual instructions with the objective of selling/ buying at a higher/ lower price. 
We set the deal price to $[\$10, \$20]$ for easier evaluation, since other the deal price may vary in a wide range according to the observations from our preliminary experiments.
To achieve this, we hard code the seller to kick off the negotiation with ``This is a good balloon and its price is \$20.''
Similarly, the buyer always opens with ``Would you consider selling it for \$10?''
When both players strictly follow the game rules, the deal price would be between \$10 and \$20.
We let the models play multiple runs and measure the average deal price before and after AI feedback. 
During the game, the seller's output is used to prompt the buyer and vice versa, conditioning on the entire conversation history.
This process is repeated till a terminal state is reached.
Fig.~\ref{fig:method:do_bargain}A shows an example round.
We define three game states:
(1) ON-GOING: the negotiation between the two players is still ongoing;
(2) DEAL: the negotiation has concluded and the two players have reached a deal; 
(3) NO DEAL: the players cannot agree on a price and have failed to reach a deal. 
To track the game states, we set an additional moderator 
(powered by a fourth LLM, in our case, \texttt{gpt-3.5-turbo}) to read the current dialog and classify the states (we will discuss more details about the moderator later). 
We measure the performance of the players based on the final deal price.

\textbf{Critic}\quad\quad 
A round is finished when the negotiation reaches a terminating state, either a DEAL or NO DEAL.
After each round, the critic LM is asked to provide constructive feedback to the player we aim to improve.
This player's dialog history from all past rounds and all feedback it has received are used to prompt the critic LM  (Fig.~\ref{fig:method:do_bargain}B).
The critic model is instructed to provide three suggestions to the player, in order to improve its negotiation strategies to achieve a more favorable price in the next game.
Before the next round, the player being improved receives the critic's feedback as a textual prompt, while its rival clears its negotiation history and restarts.

\textbf{The Moderator}\quad\quad 
The game state is classified by prompting a \texttt{gpt-3.5-turbo} moderator using few-shot demonstrations. 
The moderator reads the most recent four rounds (as well as in-context examples of different dialog states) and determines the state of the negotiation. 
Empirically, we found that four rounds of conversations are sufficient for the moderator to determine the negotiation state.
One key challenge here is detecting no-deals as the model seems to be better at recognizing DEAL than NO DEAL.
We mitigate this issue by playing multiple runs, inspect failure cases manually, and add them to the prompt with corrected labels. 
We find this method an effective side product recommend it as a technique for prompt optimization for generic classification tasks.


\textbf{Playing for Multiple Rounds}\quad\quad
Finally, we would like to explore whether the players can continuously improve from AI feedback in a game over multiple rounds.
Intuitively, the more rounds the players play, the more challenging to keep improving because the (already improved) price from the previous round becomes the baseline for the next round. 
In the experiments, we will show that only \texttt{gpt-4} can improve over 5 rounds while other models' improvements may saturate at about 3 rounds. 


\section{Related Work}
\label{sec:related_work}
\textbf{Game Playing and AlphaGo Zero}\quad\quad
Our setting is strongly inspired by AlphaGo Zero~\citep{silver2017mastering} where two agents play the game of Go and improve each other with minimal human intervention.
Here we would like to explore its counterpart in natural language. 
Our work is similar to AlphaGo Zero in the sense that we also have AI agents (large language models) playing \textit{competitive} games (bargaining) and try to improve with little human supervision. 
Yet there is an important difference between our work and AlphaGo Zero: we have a third agent, \textit{the critic}, to give feedback helping its player to improve. This is a \textit{cooperative} relationship that does not exist in AlphaGo Zero.
On the NLP side, the closest related work is~\citet{lewis2017deal} where they have (small) RNN~\citep{chung2014empirical} language models to bargain, and our work can be viewed as a more developed version of them since we change the engine to be large language models.
In general, our work is broadly under the area of AI negotiation~\citep{chawla2021casino, chawla2021casino}, strategic reasoning~\citep{meta2022human}, and general game playing~\citep{silver2016mastering}. 

\textbf{Large Language Models as Generative Agents}\quad\quad
Large language models have demonstrated incredible multi-dimensional capabilities~\citep{wei2022emergent, OpenAI2023GPT4TR}, especially in complex reasoning~\citep{wei2022chain, qiao2022reasoning, fu2022complexity} and multi-round dialog~\citep{glaese2022improving, askell2021general, bai2022constitutional}, which serve as the foundation of this work. 
Our work is related to concurrent works like Generative Agents~\citep{park2023generative} and CAMEL~\citep{li2023camel} as they also study the behavior of LLMs in a multi-agent game setting. 
The core difference between our work and theirs is that we have a clear objective (the deal price) for the model to improve through competition and cooperation, while their work studies the generic social behavior of LLMs.

\textbf{Learning from AI Feedback}\quad\quad
Our method is also strongly inspired by constitutional AI~\citep{bai2022constitutional} as we both use AI feedback, while the difference is that our feedback is directly in natural language (not a scalar from a reward model). 
There are also related/ concurrent works demonstrating the effectiveness of natural language feedback~\citep{scheurer2022training, perez2022discovering, liu2023languages} and self-refinement~\citep{chen2023teaching, madaan2023self}. 
Our work further confirms the effectiveness of AI feedback in the strategic negotiation game setting.




\section{Experiments}
\label{sec:experiments}

In our experiments, we consider three stages that gradually deepen our exploration of learning from AI feedback:
(1) We first set up the basics of the game (Sec.~\ref{ssec:basic_exps}), showing that only a few models can improve from AI critics, in which case AI feedback can be comparable (but more scalable) as human feedback.
Other models either do not understand/ follow the rule of bargaining, or cannot incorporate AI feedback for improvements. 
(2) Then we study the models’ behaviors when playing different roles (Sec.~\ref{ssec:llm_backend}).
We discover the intriguing result that buyers are in general harder to improve than sellers. 
(3) Finally, we study whether models can continuously improve over multiple rounds (Sec.~\ref{ssec:multiround_improvements}), and show a tradeoff of deal price versus success rate: although some models can continuously improve the deal price, it comes with a higher risk of breaking a deal. 
We further show evidence of negotiation in a more strategic way: both \texttt{gpt-4} and \texttt{claude-v1.3}'s responses become longer after multiple rounds of AI feedback (note that verbosity is a straightforward negotiation strategy), yet \texttt{gpt-4} is less verbose than \texttt{claude-v1.3} but achieves higher deal price and deal rate, meaning that its responses, although using fewer words, are more strategic and effective. 


\subsection{Experiment Setup}
\label{ssec:exp_setup}
\begin{figure*}[!t]
\small
  \centering
  \includegraphics[width=\linewidth]{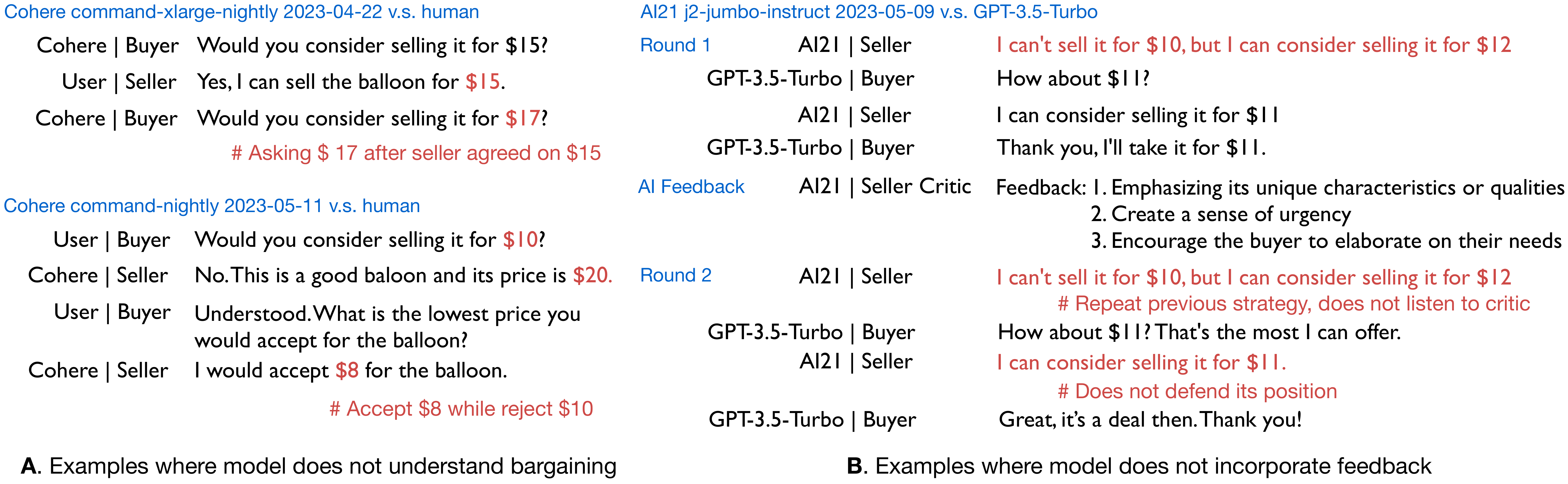}
  \caption{ \label{fig:exp:not_do_bargain}
  Not all models can play bargaining. 
  \textbf{A}. As of May 2023, the cohere model does not understand the rule of bargaining and agrees on irrational prices. 
  \textbf{B}. The AI21 Jurrasic-2 model, although understanding 
 the rule of bargaining, does not incorporate the feedback from the critic. 
 Since these models are consistently being updated, we include the timestamp and note future versions may have improved performance. 
}
\end{figure*}

\textbf{Model Engines}\quad\quad 
The minimum requirement for models to enter our game is that they should be a chatbot. 
All models we consider (\texttt{cohere-command}, AI21's \texttt{jurassic-2}, OpenAI's \texttt{gpt} and Anthropic's \texttt{claude}) can be accessed by API calls. 
Amoung them, \texttt{gpt-4} is the most expensive one and running 500 rounds of negotiation costs about \$120 and \texttt{gpt-3.5-turbo} costs about \$10. 
Other models are beta testing (as of May 2023) and do not charge money. 
For reference, the approximate rank of these models, from benchmarks like chain-of-thought hub~\citep{chain-of-thought-hub} and HeLM~\citep{liang2022holistic}, is that \texttt{gpt-4} and \texttt{claude-v1.3} are approximately similar, better than \texttt{gpt-3.5-turbo} and \texttt{claude-instant-v1.0}, and better than \texttt{cohere-command} and \texttt{j2-jumbo-instruct}. 
We will consider more models in the future, such as Google's PaLM-2~\citep{PaLM2}.


We let all models compete with \texttt{gpt-3.5-turbo}, effectively making it a baseline for all other models. 
We will show that, aligning with other concurrent model rankings~\citep{chain-of-thought-hub, liang2022holistic}, \texttt{gpt-3.5-turbo} is a middle-level powerful engine (worse than \texttt{gpt-4}, better than \texttt{claude-instant-v1.0}).
For a given model engine (say \texttt{claude-v1.3}),
we run it as the seller (with \texttt{gpt-3.5-turbo} as the buyer) and as a buyer (with \texttt{gpt-3.5-turbo} now as the seller)
We first let the models to play one round and manually inspect if they understand the rules of bargaining.
If they do,
we let them play two rounds to see if they could respond to AI feedback.
For the critic model, we set its engine the same as its player. 
We repeat the game 500 times to compute the average deal price before and after AI feedback. 
If they do improve one round, 
we let them play multiple rounds and see if they could continuously improve their strategy.
We repeat the game 200 times with 5 max rounds to compute the average deal price for each round.
When decoding from the model engines, we use sampling with default temperature (1.0 for \texttt{gpt} and \texttt{claude}, 0.75 for \texttt{cohere} and 0.7 for \texttt{j2}).


\begin{table*}[t]
  \caption{
  \textbf{Seller performance} using AI feedback vs. randomly selected human feedback from a pre-defined pool. 
  Recall that the buyer is fixed to be \texttt{gpt-3.5-turbo} and has no access to previous rounds. 
  AI's feedback is comparable to human's, but is more scalable,
  as the two both induce similar price increases. 
  }
  \label{tab:exp:human_baseline}
  \centering
  \begin{tabular}{@{}l|ccc@{}}
      \toprule
       &  \bf GPT-3.5-Turbo &\bf  Claude-instant-v1.0 &\bf Claude-v1.3\\\midrule
      Before feedback  & 16.26  & 14.74  & 15.40 \\
      Random sampled human feedback & 16.83 (\blue{+0.57}) & 16.33 (\blue{+1.59}) & 16.89 (\blue{+1.49})\\ 
      AI feedback & 17.03 (\blue{+0.77}) & 15.98 (\blue{+1.24}) & 16.98 (\blue{+1.58})\\ 
      \bottomrule
  \end{tabular}
\end{table*}

\textbf{Prompt Engineering}\quad\quad
In this work, we only had to manually optimize the prompts for the moderator  
because the player may reach/ break a deal with very diverse expressions, and we would like to make sure the moderator correctly recognizes all of them.
As mentioned above, 
we identify the errors made by the moderator in identifying deals and keep adding them as in-context demonstrations until the model reaches a sufficiently high accuracy (about 90+ by manual inspection).
For the players and the critic, we do not do prompt engineering and keep the instructions the same for all engines (but the format may be different, e.g., \texttt{claude} requires two linebreaks before ``HUMAN:'' and \texttt{j2} requires two ``\#\#'' after each dialog round).
Code and Prompts will be released publicly on publication. 


\subsection{Basic Experiments}
\label{ssec:basic_exps}

In this section, we first study the minimal requirements for models to participle in our game, namely (1) understanding the rule of bargaining and (2) responding to AI feedback. 
Then we consider basic comparison between AI and human feedback, showing that AI feedback can be comparable to human feedback, but more scalable.

\begin{figure*}[!t]
\small
  \centering
  \includegraphics[width=0.9\linewidth]{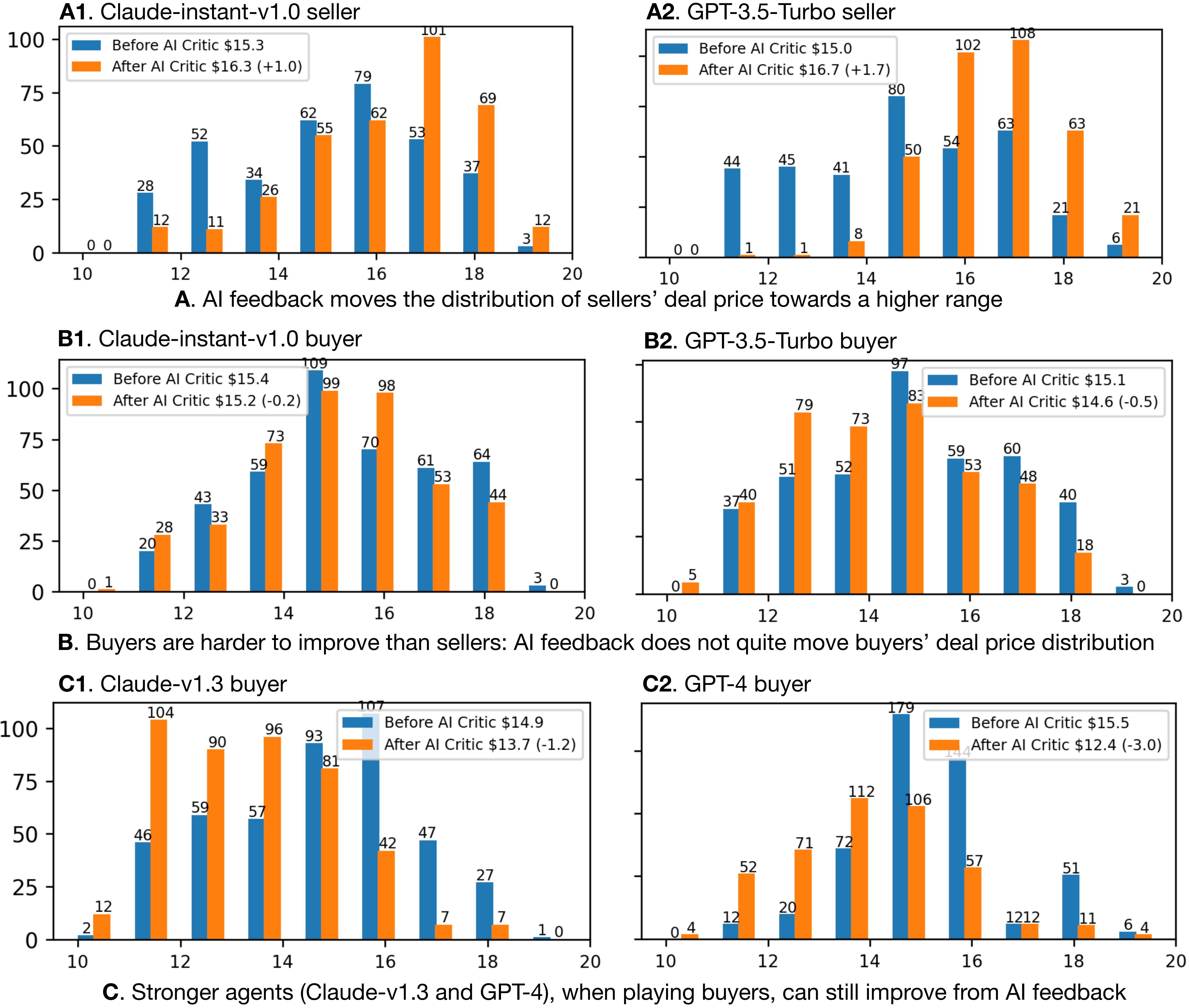}
  \caption{ \label{fig:exp:role_and_engine}
  Binned deal price frequencies of 500 games,  before v.s. after feedback.
  Effective feedback should move the distribution towards a lower/ higher price range. 
  X-axis: intervals of deals from \$10 (buyers' initial price) to \$20 (sellers' asking price). Y-axis: the frequency of the price. \textbf{A} and \textbf{B}: for weaker agents like \texttt{claude-instant-v1.0} and \texttt{gpt-3.5-turbo}, improving from AI feedback as the seller is easier than as buyer. For sellers, AI feedback moves the deal distribution to a higher range (rightward), but does not move buyers' deal distribution much. Consequently, the change in average deal price when playing as buyers (-0.2 and -0.5) is clearly smaller than those as sellers (+1.0 and +1.7) 
  \textbf{C}. Stronger agents (\texttt{claude-v1.3}/ \texttt{gpt-4}), can still improve from AI feedback even as buyers, with larger changes in average deal price (-1.2 and -3.0). 
  }
\end{figure*}

\textbf{Conversational ability does not guarantee ability to negotiate or learning from feedback}\quad\quad 
We study whether conversational models can understand the rule of bargaining by manually checking  traces of the dialog, and found that \texttt{cohere-command} fails to understand the rules, as is shown in Fig~\ref{fig:exp:not_do_bargain}A. 
We observe that it does not realize what price is a better deal. 
For example, when playing seller, it rejects a proposal of \$10 but accepts \$8.
We also observe that AI21's \texttt{j2-jumbo-instruct} model, although understanding the rule of bargaining, cannot incorporate AI feedback, as is shown in Fig.~\ref{fig:exp:not_do_bargain}B.
Generally, when instructed with AI feedback, the model keeps the same strategy as before, without any improvements. 

After ruling out the \texttt{cohere-command} and \texttt{j2-jumbo-instruct} models, we consider 
the three remaining
models: \texttt{gpt-3.5-turbo}, \texttt{claude-instant-v1.0} and \texttt{claude-v1.3}. 
For these three engines, we do not observe the problems in Fig.~\ref{fig:exp:not_do_bargain}.
This means that these models can be used for our multi-round games.

\textbf{AI Feedback can be comparable to human feedback}\quad\quad 
Now we consider some initial comparison between AI and human feedback. 
We emphasize that our goal is not to show which one is better -- a similar level of effectiveness would suffice our study (to see if LLMs can continuously improve through self-play and AI feedback). 
For the human feedback, we manually write done a pool of 10 suggestions.
Then we play 500 runs of the game, computing the deal price before and after feedback. 
After 500 runs, we compare the improvements after: 
(1) randomly sampling 3 suggestions from the predefined pool and 
(2) asking the AI critic to write down 3 suggestions. 
We note that this may underestimate the performance of human feedback, yet it would be unpractical to ask human to write done 3 suggestions for all 1500 runs (while AI feedback does not have this problem). 
The results are shown in Table~\ref{tab:exp:human_baseline} where we see that all three models (\texttt{gpt-3.5-turbo}, \texttt{claude-instant-v1.0} and \texttt{claude-v1.3}) exhibit comparable improvements over human and AI feedback.

\begin{figure*}[!t]
\small
  \centering
  \includegraphics[width=\linewidth]{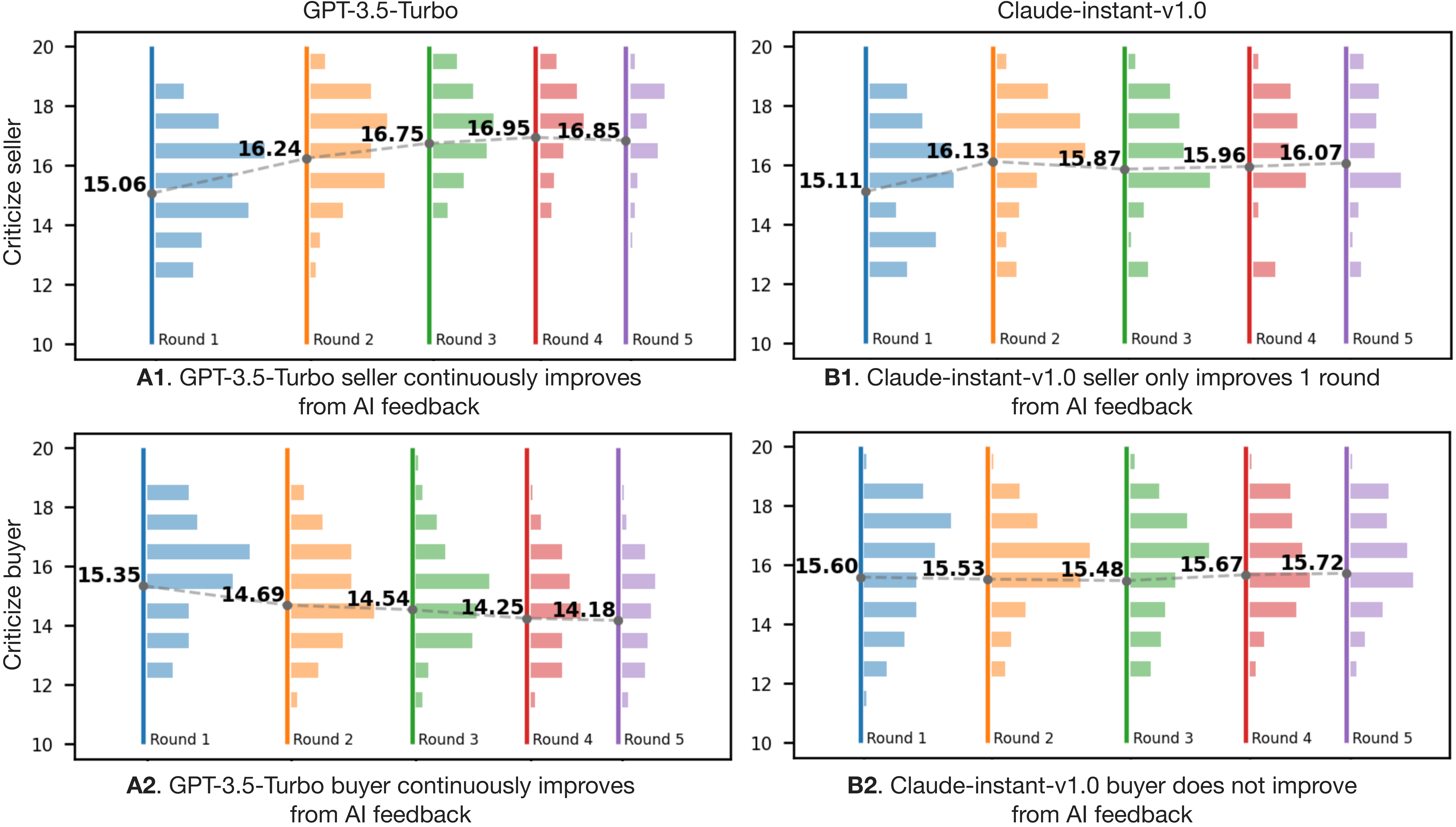}
  \caption{ \label{fig:method:multi_round}
  In the multi-round setting, different engines have different behavior when playing seller/ buyer.
  Line plots are the average price over 200 runs and bar plots represent the price distribution.
  \textbf{A1} v.s. \textbf{B1}. When playing sellers, \texttt{gpt-3.5-turbo} can improve from AI feedback in multiple rounds, while \texttt{claude-instant-v1.0} only improves the first round.  
  \textbf{A2} v.s. \textbf{B2}. When playing buyers, \texttt{gpt-3.5-turbo} can improve in multiple rounds, whild \texttt{claude-instant-v1.0} cannot. 
}
\end{figure*}

\subsection{Behaviors of Different LLM Backend}
\label{ssec:llm_backend}
So far we have established that our game setting is valid for stronger LLM engines. 
Now we consider the detailed behavior comparisons using different engines for different roles. 
Specifically, we use \texttt{claude-instant-v1.0}, \texttt{claude-v1.3}, \texttt{gpt-3.5-turbo}, and \texttt{gpt-4} to play the seller/ buyer (against a \texttt{gpt-3.5-turbo} buyer/ seller respectively), then study the deal price distribution before/ after AI feedback (also recall that the AI critic is powered by the same engine as its player). 
The results are visualized in Fig.~\ref{fig:exp:role_and_engine}. When \texttt{claude-instant-v1.0} and \texttt{gpt-3.5-turbo} play the seller, they are able to improve their average deal price after AI feedback (Fig.~\ref{fig:exp:role_and_engine}A).
But when they play the buyer role, their average deal price does not improve, which indicates that buyers tend to be a harder role than sellers (Fig.~\ref{fig:exp:role_and_engine}B). 
Yet this observation does not hold for
engines
like \texttt{gpt-4} and \texttt{claude-v1.3}, as they can still improve from AI feedback even playing buyers. 
Overall, this set of experiments reveal the nuanced capability differences between the four engines we consider. 



\subsection{Towards Continuous Improvements from Iterative AI Feedback}
\label{ssec:multiround_improvements}

Now we unroll the game to multiple rounds and see if models can continuously improve from previous dialog history and iterative AI feedback. 
Specifically, we let \texttt{gpt-3.5-turbo}, \texttt{gpt-4}, \texttt{claude-instant-v1.0}, and \texttt{claude-v1.3} play as the seller/ buyer respectively against a rival powered by \texttt{gpt-3.5-turbo}. 
As mentioned before, the critic shares the same engine as the player it helps with. 
We play 200 runs of the game, and unroll each game to be 5 rounds. 
We compute the final deal price and the deal success rate and see if the price can be continuously improved.

\begin{figure*}[!t]
\small
  \centering
  \includegraphics[width=\linewidth]{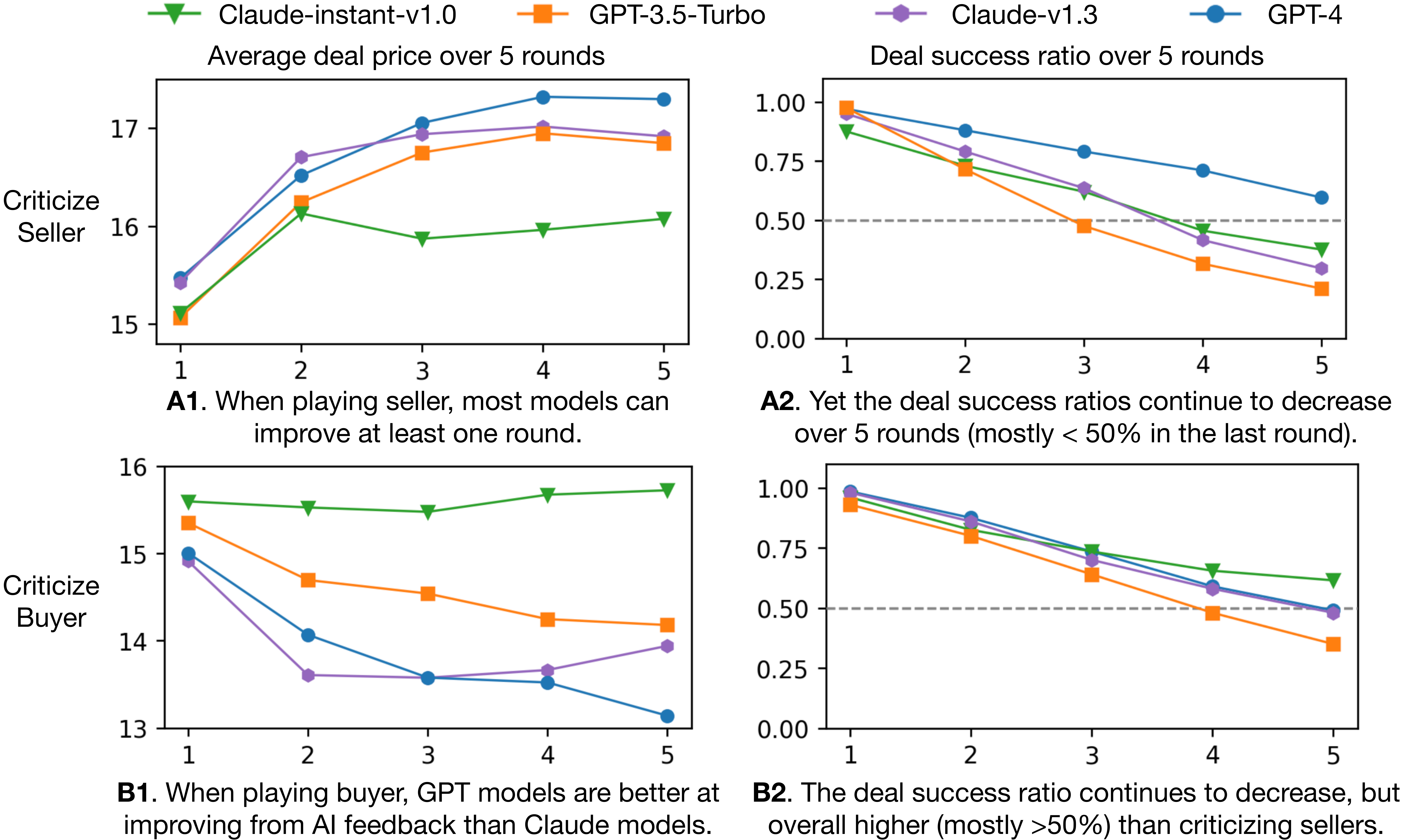}
  \caption{ \label{fig:method:multi_round_all_model}
  Performance of GPT and Claude models in multi-round games and their success rate of getting a deal. 
  \textbf{A1} and \textbf{A2}: when playing the seller, most models can improve over multiple rounds. Yet higher prices also mean that it is more likely the seller may break the deal,
  as shown in the continuously decreasing curve of A2. 
  \textbf{B1} and \textbf{B2}: when playing buyer, \texttt{claude-instant-v1.0} cannot improve over multiple rounds while others can. 
  Again, a better buying price also comes with a higher chance of 
  running away from a deal.
  We see that \texttt{GPT-4} achieves the best trade-off here: it gets the best price over multiple rounds with a higher success rate of reaching a deal. 
}
\end{figure*}

\begin{figure*}[!t]
\small
  \centering
  \includegraphics[width=\linewidth]{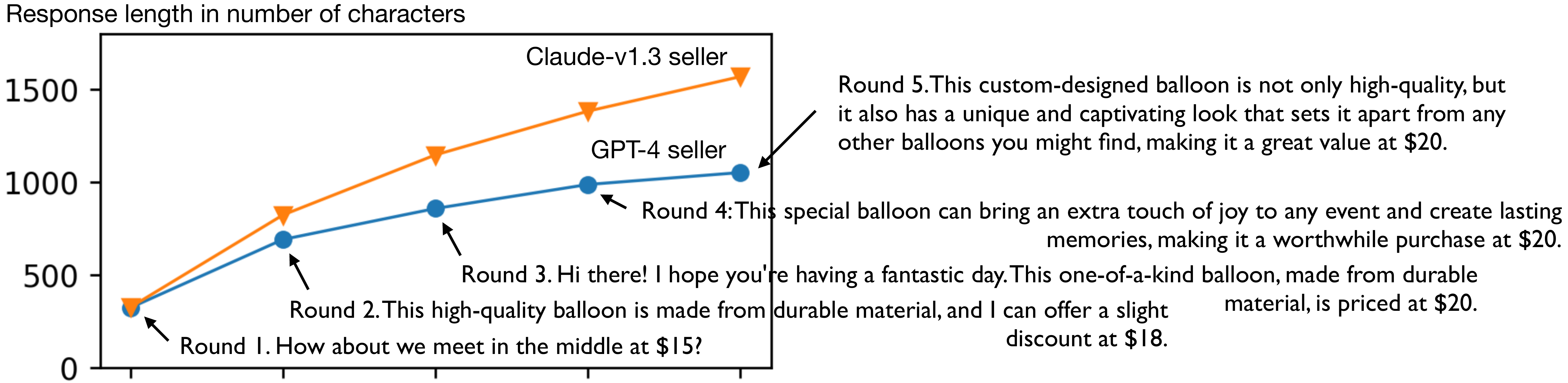}
  \caption{ \label{fig:exp:response_length}
  The average response length increases as the model learns from multiple rounds. 
  Here we show examples of the seller's response when being asked the buyer's initial query ``Would you consider selling it for \$10?'' 
  After multiple rounds of negotiation, the seller's responses become more verbose and word-tuned. 
  Yet verbosity does not mean better strategy: \texttt{claude-v1.3} is more verbose (higher curve) than \texttt{gpt-4}, but it has a worse success rate and deal price (recall Fig.~\ref{fig:method:multi_round_all_model}).
  This indicates that \texttt{gpt-4}'s verbosity is more strategic. 
  }
\end{figure*}

Fig.~\ref{fig:method:multi_round} shows  \texttt{gpt-3.5-turbo} and \texttt{claude-instant-v1.0} playing different roles. 
For a given engine, improvements over one round may not necessarily extrapolate to multiple rounds, as we observe that \texttt{gpt-3.5-turbo} can improve over multiple rounds, but \texttt{claude-instant-v1.0} only improves at most one round. 

Now we consider the tradeoff between the tendency of achieving a higher deal price versus the rick of breaking a deal, as is shown in Fig~\ref{fig:method:multi_round_all_model}. 
We see that when playing sellers, all four model engines can improve over at least one round, but this comes at the cost of decreasing deal success ratio. 
When playing buyers, there are models that cannot improve (\texttt{claude-instant-v1.0}), or saturate over 3 rounds (\texttt{claude-v1.3}), while \texttt{gpt-4} and \texttt{gpt-3.5-turbo} can continuously improve, and \texttt{gpt-4} achieves better (lower) deal price and higher deal rate than \texttt{gpt-3.5-turbo}. 

Finally, we study how iterative AI feedback influences the language complexity used by the agents by plotting the average response length (measured in number of characters) after each round, as is shown in Fig.~\ref{fig:exp:response_length}. 
We see that both \texttt{claude-v1.3} and \texttt{gpt-4} become more verbose after iterative AI feedback with a continuously increasing response length. 
This is intuitive because being verbosity is a straightforward strategy in negotiation. 
Yet for \texttt{claude-v1.3}, the verbosity does not translate to better negotiation strategy, as its improvement saturates after three rounds (Fig.~\ref{fig:method:multi_round_all_model}B1). 
In comparison, \texttt{gpt-4}'s increase verbosity is more strategic, as it use less words than \texttt{claude-v1.3}, but achieves better deal price and deal success rate (Fig.~\ref{fig:method:multi_round_all_model}B). 
This observation serve as strong evidence that AI feedback improves players' response towards a word-tuned, strategic direction.

\section{Conclusions}
\label{sec:conclusion}
In this work, we study whether multiple large language models can autonomously improve each other in a negotiation game by role-playing and learning from AI feedback. 
Our experiments show that certain
models can indeed improve by
continuously playing competition games with iterative AI feedback, under well-defined rules in an AlphaGo Zero fashion. 
We also show the tradeoff between next-round price improvement and success rate, as a better deal price also comes with a higher risk of deal breaking. 
This suggests future research may consider global optimization for improving the overall gain over multiple rounds. 
We further show evidence of improved language from iterative AI feedback: in a multi-round game, one model may be less verbose than another, but be better word-tuned, thus more effective in getting a better deal.

We believe our results have profound implications for AI research:
on the positive side, it indicates the possibility of continuously improving language models with minimal human intervention.
On the risky side, it might be more challenging to oversight the model behavior in our framework because models are acting autonomously, 
which calls for future alignment and safety research in the multi-agent game setting. 
Overall, we believe our work provides a meaningful initial exploration for large language models' learning from game-playing and iterative AI feedback.


\bibliographystyle{plainnat}
\bibliography{gpt_bargaining} 

\begin{thebibliography}{34}
\providecommand{\natexlab}[1]{#1}
\providecommand{\url}[1]{\texttt{#1}}
\expandafter\ifx\csname urlstyle\endcsname\relax
  \providecommand{\doi}[1]{doi: #1}\else
  \providecommand{\doi}{doi: \begingroup \urlstyle{rm}\Url}\fi

\bibitem[Aky{\"u}rek et~al.(2022)Aky{\"u}rek, Schuurmans, Andreas, Ma, and
  Zhou]{akyurek2022learning}
Ekin Aky{\"u}rek, Dale Schuurmans, Jacob Andreas, Tengyu Ma, and Denny Zhou.
\newblock What learning algorithm is in-context learning? investigations with
  linear models.
\newblock \emph{arXiv preprint arXiv:2211.15661}, 2022.

\bibitem[Askell et~al.(2021)Askell, Bai, Chen, Drain, Ganguli, Henighan, Jones,
  Joseph, Mann, DasSarma, et~al.]{askell2021general}
Amanda Askell, Yuntao Bai, Anna Chen, Dawn Drain, Deep Ganguli, Tom Henighan,
  Andy Jones, Nicholas Joseph, Ben Mann, Nova DasSarma, et~al.
\newblock A general language assistant as a laboratory for alignment.
\newblock \emph{arXiv preprint arXiv:2112.00861}, 2021.

\bibitem[Bai et~al.(2022)Bai, Kadavath, Kundu, Askell, Kernion, Jones, Chen,
  Goldie, Mirhoseini, McKinnon, et~al.]{bai2022constitutional}
Yuntao Bai, Saurav Kadavath, Sandipan Kundu, Amanda Askell, Jackson Kernion,
  Andy Jones, Anna Chen, Anna Goldie, Azalia Mirhoseini, Cameron McKinnon,
  et~al.
\newblock Constitutional ai: Harmlessness from ai feedback.
\newblock \emph{arXiv preprint arXiv:2212.08073}, 2022.

\bibitem[Bowman et~al.(2022)Bowman, Hyun, Perez, Chen, Pettit, Heiner,
  Lukosuite, Askell, Jones, Chen, et~al.]{bowman2022measuring}
Samuel~R Bowman, Jeeyoon Hyun, Ethan Perez, Edwin Chen, Craig Pettit, Scott
  Heiner, Kamile Lukosuite, Amanda Askell, Andy Jones, Anna Chen, et~al.
\newblock Measuring progress on scalable oversight for large language models.
\newblock \emph{arXiv preprint arXiv:2211.03540}, 2022.

\bibitem[Brown et~al.(2020)Brown, Mann, Ryder, Subbiah, Kaplan, Dhariwal,
  Neelakantan, Shyam, Sastry, Askell, et~al.]{brown2020language}
Tom Brown, Benjamin Mann, Nick Ryder, Melanie Subbiah, Jared~D Kaplan, Prafulla
  Dhariwal, Arvind Neelakantan, Pranav Shyam, Girish Sastry, Amanda Askell,
  et~al.
\newblock Language models are few-shot learners.
\newblock \emph{Advances in neural information processing systems},
  33:\penalty0 1877--1901, 2020.

\bibitem[Chawla et~al.(2021)Chawla, Ramirez, Clever, Lucas, May, and
  Gratch]{chawla2021casino}
Kushal Chawla, Jaysa Ramirez, Rene Clever, Gale Lucas, Jonathan May, and
  Jonathan Gratch.
\newblock Casino: A corpus of campsite negotiation dialogues for automatic
  negotiation systems.
\newblock \emph{arXiv preprint arXiv:2103.15721}, 2021.

\bibitem[Chen et~al.(2023)Chen, Lin, Sch{\"a}rli, and Zhou]{chen2023teaching}
Xinyun Chen, Maxwell Lin, Nathanael Sch{\"a}rli, and Denny Zhou.
\newblock Teaching large language models to self-debug.
\newblock \emph{arXiv preprint arXiv:2304.05128}, 2023.

\bibitem[Chowdhery et~al.(2022)Chowdhery, Narang, Devlin, Bosma, Mishra,
  Roberts, Barham, Chung, Sutton, Gehrmann, et~al.]{chowdhery2022palm}
Aakanksha Chowdhery, Sharan Narang, Jacob Devlin, Maarten Bosma, Gaurav Mishra,
  Adam Roberts, Paul Barham, Hyung~Won Chung, Charles Sutton, Sebastian
  Gehrmann, et~al.
\newblock Palm: Scaling language modeling with pathways.
\newblock \emph{arXiv preprint arXiv:2204.02311}, 2022.

\bibitem[Chung et~al.(2014)Chung, Gulcehre, Cho, and
  Bengio]{chung2014empirical}
Junyoung Chung, Caglar Gulcehre, KyungHyun Cho, and Yoshua Bengio.
\newblock Empirical evaluation of gated recurrent neural networks on sequence
  modeling.
\newblock \emph{arXiv preprint arXiv:1412.3555}, 2014.

\bibitem[Cohere(2023)]{cohere}
Cohere.
\newblock Cohere command models.
\newblock \emph{Cohere website}, 2023.
\newblock URL \url{https://docs.cohere.com/docs/models}.

\bibitem[Dai et~al.(2022)Dai, Sun, Dong, Hao, Sui, and Wei]{dai2022can}
Damai Dai, Yutao Sun, Li~Dong, Yaru Hao, Zhifang Sui, and Furu Wei.
\newblock Why can gpt learn in-context? language models secretly perform
  gradient descent as meta optimizers.
\newblock \emph{arXiv preprint arXiv:2212.10559}, 2022.

\bibitem[(FAIR)† et~al.(2022)(FAIR)†, Bakhtin, Brown, Dinan, Farina,
  Flaherty, Fried, Goff, Gray, Hu, et~al.]{meta2022human}
Meta Fundamental AI Research Diplomacy~Team (FAIR)†, Anton Bakhtin, Noam
  Brown, Emily Dinan, Gabriele Farina, Colin Flaherty, Daniel Fried, Andrew
  Goff, Jonathan Gray, Hengyuan Hu, et~al.
\newblock Human-level play in the game of diplomacy by combining language
  models with strategic reasoning.
\newblock \emph{Science}, 378\penalty0 (6624):\penalty0 1067--1074, 2022.

\bibitem[Fu et~al.(2022)Fu, Peng, Sabharwal, Clark, and Khot]{fu2022complexity}
Yao Fu, Hao Peng, Ashish Sabharwal, Peter Clark, and Tushar Khot.
\newblock Complexity-based prompting for multi-step reasoning.
\newblock \emph{arXiv preprint arXiv:2210.00720}, 2022.

\bibitem[Fu et~al.(2023)Fu, Ou, Chen, and Wan]{chain-of-thought-hub}
Yao Fu, Litu Ou, Mingyu Chen, and Yuhao Wan.
\newblock Measuring llms' reasoning performance.
\newblock \emph{Github}, 2023.
\newblock URL \url{https://github.com/FranxYao/chain-of-thought-hub}.

\bibitem[Glaese et~al.(2022)Glaese, McAleese, Tr{\k{e}}bacz, Aslanides, Firoiu,
  Ewalds, Rauh, Weidinger, Chadwick, Thacker, et~al.]{glaese2022improving}
Amelia Glaese, Nat McAleese, Maja Tr{\k{e}}bacz, John Aslanides, Vlad Firoiu,
  Timo Ewalds, Maribeth Rauh, Laura Weidinger, Martin Chadwick, Phoebe Thacker,
  et~al.
\newblock Improving alignment of dialogue agents via targeted human judgements.
\newblock \emph{arXiv preprint arXiv:2209.14375}, 2022.

\bibitem[Google(2023)]{PaLM2}
Google.
\newblock Palm 2 technical report.
\newblock \emph{ArXiv}, 2023.

\bibitem[Hoffmann et~al.(2022)Hoffmann, Borgeaud, Mensch, Buchatskaya, Cai,
  Rutherford, Casas, Hendricks, Welbl, Clark, et~al.]{hoffmann2022training}
Jordan Hoffmann, Sebastian Borgeaud, Arthur Mensch, Elena Buchatskaya, Trevor
  Cai, Eliza Rutherford, Diego de~Las Casas, Lisa~Anne Hendricks, Johannes
  Welbl, Aidan Clark, et~al.
\newblock Training compute-optimal large language models.
\newblock \emph{arXiv preprint arXiv:2203.15556}, 2022.

\bibitem[Labs(2023)]{jurassic}
AI21 Labs.
\newblock Announcing jurassic-2 and task-specific apis.
\newblock \emph{AI21 Blog}, 2023.
\newblock URL \url{https://www.ai21.com/blog/introducing-j2}.

\bibitem[Lewis et~al.(2017)Lewis, Yarats, Dauphin, Parikh, and
  Batra]{lewis2017deal}
Mike Lewis, Denis Yarats, Yann~N Dauphin, Devi Parikh, and Dhruv Batra.
\newblock Deal or no deal? end-to-end learning for negotiation dialogues.
\newblock \emph{arXiv preprint arXiv:1706.05125}, 2017.

\bibitem[Li et~al.(2023)Li, Hammoud, Itani, Khizbullin, and
  Ghanem]{li2023camel}
Guohao Li, Hasan Abed Al~Kader Hammoud, Hani Itani, Dmitrii Khizbullin, and
  Bernard Ghanem.
\newblock Camel: Communicative agents for" mind" exploration of large scale
  language model society.
\newblock \emph{arXiv preprint arXiv:2303.17760}, 2023.

\bibitem[Liang et~al.(2022)Liang, Bommasani, Lee, Tsipras, Soylu, Yasunaga,
  Zhang, Narayanan, Wu, Kumar, et~al.]{liang2022holistic}
Percy Liang, Rishi Bommasani, Tony Lee, Dimitris Tsipras, Dilara Soylu,
  Michihiro Yasunaga, Yian Zhang, Deepak Narayanan, Yuhuai Wu, Ananya Kumar,
  et~al.
\newblock Holistic evaluation of language models.
\newblock \emph{arXiv preprint arXiv:2211.09110}, 2022.

\bibitem[Liu et~al.(2023)Liu, Sferrazza, and Abbeel]{liu2023languages}
Hao Liu, Carmelo Sferrazza, and Pieter Abbeel.
\newblock Languages are rewards: Hindsight finetuning using human feedback.
\newblock \emph{arXiv preprint arXiv:2302.02676}, 2023.

\bibitem[Madaan et~al.(2023)Madaan, Tandon, Gupta, Hallinan, Gao, Wiegreffe,
  Alon, Dziri, Prabhumoye, Yang, et~al.]{madaan2023self}
Aman Madaan, Niket Tandon, Prakhar Gupta, Skyler Hallinan, Luyu Gao, Sarah
  Wiegreffe, Uri Alon, Nouha Dziri, Shrimai Prabhumoye, Yiming Yang, et~al.
\newblock Self-refine: Iterative refinement with self-feedback.
\newblock \emph{arXiv preprint arXiv:2303.17651}, 2023.

\bibitem[OpenAI(2023)]{OpenAI2023GPT4TR}
OpenAI.
\newblock Gpt-4 technical report.
\newblock \emph{ArXiv}, abs/2303.08774, 2023.

\bibitem[Ouyang et~al.(2022)Ouyang, Wu, Jiang, Almeida, Wainwright, Mishkin,
  Zhang, Agarwal, Slama, Ray, et~al.]{ouyang2022training}
Long Ouyang, Jeffrey Wu, Xu~Jiang, Diogo Almeida, Carroll Wainwright, Pamela
  Mishkin, Chong Zhang, Sandhini Agarwal, Katarina Slama, Alex Ray, et~al.
\newblock Training language models to follow instructions with human feedback.
\newblock \emph{Advances in Neural Information Processing Systems},
  35:\penalty0 27730--27744, 2022.

\bibitem[Park et~al.(2023)Park, O'Brien, Cai, Morris, Liang, and
  Bernstein]{park2023generative}
Joon~Sung Park, Joseph~C O'Brien, Carrie~J Cai, Meredith~Ringel Morris, Percy
  Liang, and Michael~S Bernstein.
\newblock Generative agents: Interactive simulacra of human behavior.
\newblock \emph{arXiv preprint arXiv:2304.03442}, 2023.

\bibitem[Perez et~al.(2022)Perez, Ringer, Luko{\v{s}}i{\=u}t{\.e}, Nguyen,
  Chen, Heiner, Pettit, Olsson, Kundu, Kadavath, et~al.]{perez2022discovering}
Ethan Perez, Sam Ringer, Kamil{\.e} Luko{\v{s}}i{\=u}t{\.e}, Karina Nguyen,
  Edwin Chen, Scott Heiner, Craig Pettit, Catherine Olsson, Sandipan Kundu,
  Saurav Kadavath, et~al.
\newblock Discovering language model behaviors with model-written evaluations.
\newblock \emph{arXiv preprint arXiv:2212.09251}, 2022.

\bibitem[Qiao et~al.(2022)Qiao, Ou, Zhang, Chen, Yao, Deng, Tan, Huang, and
  Chen]{qiao2022reasoning}
Shuofei Qiao, Yixin Ou, Ningyu Zhang, Xiang Chen, Yunzhi Yao, Shumin Deng,
  Chuanqi Tan, Fei Huang, and Huajun Chen.
\newblock Reasoning with language model prompting: A survey.
\newblock \emph{arXiv preprint arXiv:2212.09597}, 2022.

\bibitem[Scheurer et~al.(2022)Scheurer, Campos, Chan, Chen, Cho, and
  Perez]{scheurer2022training}
J{\'e}r{\'e}my Scheurer, Jon~Ander Campos, Jun~Shern Chan, Angelica Chen,
  Kyunghyun Cho, and Ethan Perez.
\newblock Training language models with natural language feedback.
\newblock \emph{arXiv preprint arXiv:2204.14146}, 2022.

\bibitem[Silver et~al.(2016)Silver, Huang, Maddison, Guez, Sifre, Van
  Den~Driessche, Schrittwieser, Antonoglou, Panneershelvam, Lanctot,
  et~al.]{silver2016mastering}
David Silver, Aja Huang, Chris~J Maddison, Arthur Guez, Laurent Sifre, George
  Van Den~Driessche, Julian Schrittwieser, Ioannis Antonoglou, Veda
  Panneershelvam, Marc Lanctot, et~al.
\newblock Mastering the game of go with deep neural networks and tree search.
\newblock \emph{nature}, 529\penalty0 (7587):\penalty0 484--489, 2016.

\bibitem[Silver et~al.(2017)Silver, Schrittwieser, Simonyan, Antonoglou, Huang,
  Guez, Hubert, Baker, Lai, Bolton, et~al.]{silver2017mastering}
David Silver, Julian Schrittwieser, Karen Simonyan, Ioannis Antonoglou, Aja
  Huang, Arthur Guez, Thomas Hubert, Lucas Baker, Matthew Lai, Adrian Bolton,
  et~al.
\newblock Mastering the game of go without human knowledge.
\newblock \emph{nature}, 550\penalty0 (7676):\penalty0 354--359, 2017.

\bibitem[von Oswald et~al.(2022)von Oswald, Niklasson, Randazzo, Sacramento,
  Mordvintsev, Zhmoginov, and Vladymyrov]{von2022transformers}
Johannes von Oswald, Eyvind Niklasson, Ettore Randazzo, Jo{\~a}o Sacramento,
  Alexander Mordvintsev, Andrey Zhmoginov, and Max Vladymyrov.
\newblock Transformers learn in-context by gradient descent.
\newblock \emph{arXiv preprint arXiv:2212.07677}, 2022.

\bibitem[Wei et~al.(2022{\natexlab{a}})Wei, Tay, Bommasani, Raffel, Zoph,
  Borgeaud, Yogatama, Bosma, Zhou, Metzler, et~al.]{wei2022emergent}
Jason Wei, Yi~Tay, Rishi Bommasani, Colin Raffel, Barret Zoph, Sebastian
  Borgeaud, Dani Yogatama, Maarten Bosma, Denny Zhou, Donald Metzler, et~al.
\newblock Emergent abilities of large language models.
\newblock \emph{arXiv preprint arXiv:2206.07682}, 2022{\natexlab{a}}.

\bibitem[Wei et~al.(2022{\natexlab{b}})Wei, Wang, Schuurmans, Bosma, Chi, Le,
  and Zhou]{wei2022chain}
Jason Wei, Xuezhi Wang, Dale Schuurmans, Maarten Bosma, Ed~Chi, Quoc Le, and
  Denny Zhou.
\newblock Chain of thought prompting elicits reasoning in large language
  models.
\newblock \emph{arXiv preprint arXiv:2201.11903}, 2022{\natexlab{b}}.

\end{thebibliography}


\end{document}